\documentclass{article}

\usepackage{microtype}
\usepackage{graphicx}
\usepackage{subfigure}
\usepackage{booktabs} 

\usepackage{epsfig}
\usepackage{graphicx}
\usepackage{hyperref}
\usepackage{url}
\usepackage[usenames, dvipsnames]{color}
\usepackage[super]{nth}

\newcommand{\Gaussian}{\mathcal{N}}

\def\-{\raisebox{0.5pt}{-}}

\usepackage{hyperref}

\usepackage[accepted]{icml2019}

\icmltitlerunning{Accelerating Training of Deep Neural Networks with a Standardization Loss}

\begin{document}

\twocolumn[
\icmltitle{Accelerating Training of Deep Neural Networks with a Standardization Loss}
\icmlsetsymbol{equal}{*}

\begin{icmlauthorlist}
\icmlauthor{Jasmine Collins}{berk}
\icmlauthor{Johannes Ball\'e}{goog}
\icmlauthor{Jonathon Shlens}{goog}
\end{icmlauthorlist}

\icmlaffiliation{berk}{University of California, Berkeley}
\icmlaffiliation{goog}{Google AI}

\icmlcorrespondingauthor{Jasmine Collins}{jazzie@berkeley.edu}

\vskip 0.3in
]

\printAffiliationsAndNotice{}

\begin{abstract}
A significant advance in accelerating neural network training has been the development of normalization methods \citep{BatchNorm, wu2018group, ba2016layer, salimans2016weight}, permitting the training of deep models both faster and with better accuracy.
These advances come with practical challenges: for instance, batch normalization ties the prediction of individual examples with other examples within a batch, resulting in a network that is heavily dependent on batch size. Layer normalization and group normalization are data-dependent and thus must be continually used, even at test-time. 
To address the issues that arise from using explicit normalization techniques, 
we propose to replace existing normalization methods with a simple, secondary objective loss that we term a {\it standardization loss}. 
This formulation is flexible and robust across different batch sizes and surprisingly, this secondary objective accelerates learning on the primary training objective. Because it is a training loss, it is simply removed at test-time, and no further effort is needed to maintain normalized activations.
We find that a standardization loss accelerates training on both small- and large-scale image classification experiments, works with a variety of architectures, and is largely robust to training across different batch sizes. 
\end{abstract}
\section{Introduction}

Recent progress in machine learning has been fueled by the recognition that a key ingredient of state-of-the-art systems is building larger models on increasingly larger datasets \citep{halevy2009unreasonable}. Deep learning \citep{lecun2015deep, goodfellow2016deep} has been one such direction where models provide increasing gains in predictive performance as model size and datasets grow (e.g. \cite{williams2015scaling}). The limiting factor in building deep learning models is thus the speed at which one may train such systems on large amounts of data. This limitation has yielded new classes of specialized hardware \citep{jouppi2018motivation,chetlur2014cudnn} as well as driven the development of new model architectures \citep{ioffe2017batch, wu2018group, lstm} and optimization methods \citep{duchi2011adaptive,Tieleman2012rmsprop, ADAM} -- all with the goal of accelerating neural network training.

Batch normalization (BN) \citep{BatchNorm} is one method for accelerating neural network training that has become a necessary and standard component in many state-of-the-art systems in image classification \citep{szegedy2016rethinking, szegedy2016inception, he2015deep, identity-mappings}, object detection \citep{huang2016speed}, image segmentation \citep{chen2018deeplab,zhao2017pyramid}, generative models \citep{radford2015unsupervised,karras2017progressive} and other problems \citep{cooijmans2016recurrent}. BN explicitly requires a network to maintain normalized (i.e. standardized) activations. The resulting network benefits from accelerated learning by up to an order-of-magnitude \citep{BatchNorm}, and makes some networks trainable that were previously untrainable (e.g. \citet{radford2015unsupervised}).

In spite of its successes, BN does offer some serious challenges and limitations: (1) BN is poorly understood and there is no general consensus on how it works \citep{santurkar2018does,bjorck2018understanding,kohler2018towards} making a successful application of BN to some architectures challenging \citep{ba2016layer, laurent2016batch, cooijmans2016recurrent}, (2) BN complicates training significantly because it couples individual training examples and makes training on small batch sizes or non {\it i.i.d.} statistics difficult (but see \cite{ioffe2017batch}), (3) BN does not permit a unified training objective because some parameters (i.e. moments used for inference) are estimated with moving averages independent of the training objective.

Alternative forms of normalization have since been proposed that act on different dimensions of the network activations. Layer normalization \cite{ba2016layer} and group normalization \cite{wu2018group} both reduce the dependency of normalized models on batch dimension, but introduce the drawback that they require continued use at test-time. Because BN uses separately estimated normalization statistics at inference, these parameters may be 'folded in' to the previous layer's weights, ensuring that no extra operations are spent on normalization during inference. Layer and group normalization are data-dependent, and thus must be continually used at test-time.

In this work, we propose a simple method for accelerating neural network training that achieves some of the gains of existing explicit normalization techniques without the need to re-tune hyperparameters. Instead of imposing normalization as a requirement in a network's structure, we follow a strategy of expressing normalization as an objective loss. 
This leads to a flexible form of normalization that is robust across different batch sizes for stochastic gradient descent (SGD) and simplifies behavior during inference. Surprisingly, even though the network is now required to minimize a secondary loss, we find that the resulting network trains faster on the original task. Because our approach only involves a loss (rather than making an architectural change), inference-time merely requires removing the loss -- as is naturally done after training any machine learning system. In particular, we provide the following contributions in this work:
\begin{enumerate}
    \item Introduce a secondary objective to penalize activation patterns that diverge from standardized Gaussian distributions.
    \item Demonstrate that applying this loss in a small multi-layer perceptron (MLP) and convolutional neural network (CNN) accelerates early training and is somewhat robust to a single added hyperparameter.
    \item Demonstrate that such a loss accelerates training up to 2.8\,x on large-scale MLPs and a wide range of CNN architectures trained on ImageNet.
    \item Show that training with a standardization loss instead of BN results in networks largely insensitive to batch size and provides competitive performance with existing methods.
\end{enumerate}
The outline of this paper is to first discuss the many related techniques in the literature and then introduce the proposed objective to promote standardization. In the following sections, we describe early experiments on a small dataset (MNIST) and demonstrate the relative benefit of this method for accelerating training. 
Next, we present several results on large-scale classification tasks on image classification \citep{deng2009imagenet} and 3-D point cloud classification \citep{wu20153d}, highlighting how training speed with the standardization loss is accelerated across a wide variety of architectures.
Finally, we show that networks trained with the standardization loss are largely insensitive to batch size.

\section{Related Work}

\subsection{Normalization for accelerating deep learning training}
Whitening transformations have long been used as a standard method for preprocessing data to accelerate statistical estimation and learning \citep{murphy2013ml}. Whitening, however, can be computationally challenging for high-dimensional distributions because it requires calculating and inverting a large covariance matrix. These problems are further exacerbated in deep learning systems in which multiple representations are learned simultaneously in an online fashion. These dual challenges make whitening largely intractable in deep learning systems (but see \citet{huang2018decorrelated}) and instead has lead to the development of weaker forms of imposing normalization during training.

BN \citep{BatchNorm} is a state-of-the-art solution which explicitly requires that activations in a given layer $\mathbf{x}$ must be standardized (i.e. skipping decorrelation). BN constructs the network representation to be self-normalized across the batch and spatial dimensions in a CNN, and only the batch dimension for an MLP. While BN can greatly accelerate training, it is widely known that this performance boost is predicated on the ability to train with sufficiently large batch size. For smaller batch sizes, the performance gain due to BN drops.

To alleviate issues with small batch sizes, batch renormalization \citep{ioffe2017batch} introduced two parameters to constrain the first two moments of the distribution. Although this technique was developed to mitigate issues with small batch size, models with batch renormalization still suffer from a decrease in accuracy when trained on very small batch size and still make training dependent on the individual items within a batch.

 Another approach that was recently proposed for image models is group normalization (GN) \citep{wu2018group}. GN is a normalization technique specialized for CNNs that divides a given layer into artificial groups, across which the mean and variance are respectively normalized. The resulting models are largely insensitive to batch size but perform below BN in a large batch setting \citep{wu2018group}. Additionally, GN makes no claim about accelerating the speed of training but rather focuses on asymptotic performance.

Finally, other approaches have focused on encouraging normalization during training by improving weight initialization \citep{kingma2018glow} or employing a factorized representation such that most network parameters are L2-normalized \citep{salimans2016weight} (see also \cite{arpit2016normalization}).

\subsection{Auxiliary losses to shape the distribution of activations}

Instead of building an architecture to constrain that a given layer's activations are standardized, we instead focus on constructing an auxiliary loss to shape the distribution of the activations. Applying a loss to shape the distribution of activations has been explored in the context of autoencoders to sparsify a given layer's representation \citep{lee2008sparse,nair20093d,gao2016group,ngiam2011sparse}. Specifically, in order to encourage sparsity, the discretized activations of a given layer were penalized with respect to their distance from a Bernoulli distribution. 

Similarly, \citet{goroshin2013saturating} 
added an auxiliary objective to networks in order to encourage the representation to be close to the saturating regime of a network's nonlinearities. The goal of this secondary objective was to improve generalization
and provide a defense against adversarial attacks \citep{gu2014towards}.

\section{Methods}

The following methods focus on training MLPs and CNNs \citep{lecun1998gradient}, but these methods may be extended to other network architectures \citep{goodfellow2016deep}. Let $\mathbf{x}$ be the activation pattern in a given layer of a network, before the nonlinearity is applied. For instance, for a CNN, $\mathbf{x}$ may be a tensor with dimensions $[b, h, w, n]$ corresponding to batch, height, width and channels. For an MLP, $\mathbf{x}$ may be a tensor with dimensions $[b, n]$ corresponding to $n$ neurons.

We consider each $\mathbf{x}$ to be a sample from a $n$-dimensional distribution $P(\mathbf{x})$. We assume that each example in a batch corresponds to a sample from $P(\mathbf{x})$ \citep{BatchNorm}. Additionally, in the case of a CNN, we assume ergodicity such that samples across spatial dimensions also represent independent samples from $P(\mathbf{x})$. Hence, in the case of a CNN, we can define $P(\mathbf{x})$ to be an $n$-dimensional distribution across channels in which the maximum likelihood estimate of the mean and variance are the averages across space and batch dimensions.

In the case of BN, one constructs the architecture to impose standardization on the first two moments of $P(\mathbf{x})$. We propose to replace the explicit normalization with a regularization loss that mimics the invariances offered by BN. Consider the KL-divergence of $P(\mathbf{x})$ with respect to a target distribution $Q(\mathbf{x})$:

\begin{equation}
D_{\scriptscriptstyle KL}\big(P(\mathbf{x})\;||
\; Q(\mathbf{x}) \big) = \int P(\mathbf{x}) \log \frac{P(\mathbf{x})}{Q(\mathbf{x})} d\mathbf{x}
\end{equation}

We define the \textit{standardization loss} as the KL-divergence of $P(\mathbf{x})$ with respect to a standardized Gaussian distribution $\Gaussian(\mathbf{x}; \mathbf{0}, \mathbf{I})$. The standardization loss measures how far a given distribution is from a standardized representation. The loss is calculated at each layer, and the total loss is the sum across all layers. We apply this loss during training by adding it to the primary objective (e.g. cross-entropy loss) with a selected weight.

We make a simplifying assumption that $P(\mathbf{x})$ is Gaussian distributed corresponding to a maximum entropy distribution constrained by the first two moments. The integral for the loss may then be analytically derived (\citet{kingma2013auto} provide a short derivation):

\begin{equation}
D_{\scriptscriptstyle KL}\big(P(\mathbf{x})\;||\;\Gaussian(\mathbf{x}; \mathbf{0}, \mathbf{I})\big) = \frac{1}{2} \sum^{n}_{i=1} (\mu_i^2 + \sigma_i^2 - \log (\sigma_i^2) - 1)
\end{equation}

where $\mu_i$ and $\sigma_i^2$ are the  mean and variance of dimension $i$ measured from $P(\mathbf{x})$. These statistics are calculated the same way they are calculated in BN, by marginalizing across the spatial and batch dimensions of $\mathbf{x}$ for a CNN or just the batch dimensions of an MLP.

\section{Results}

\subsection{Standardization loss accelerates training on MNIST\label{sec:mnist_cnn}}

\begin{figure}[t]
\begin{center}
\includegraphics[width=0.75\linewidth]{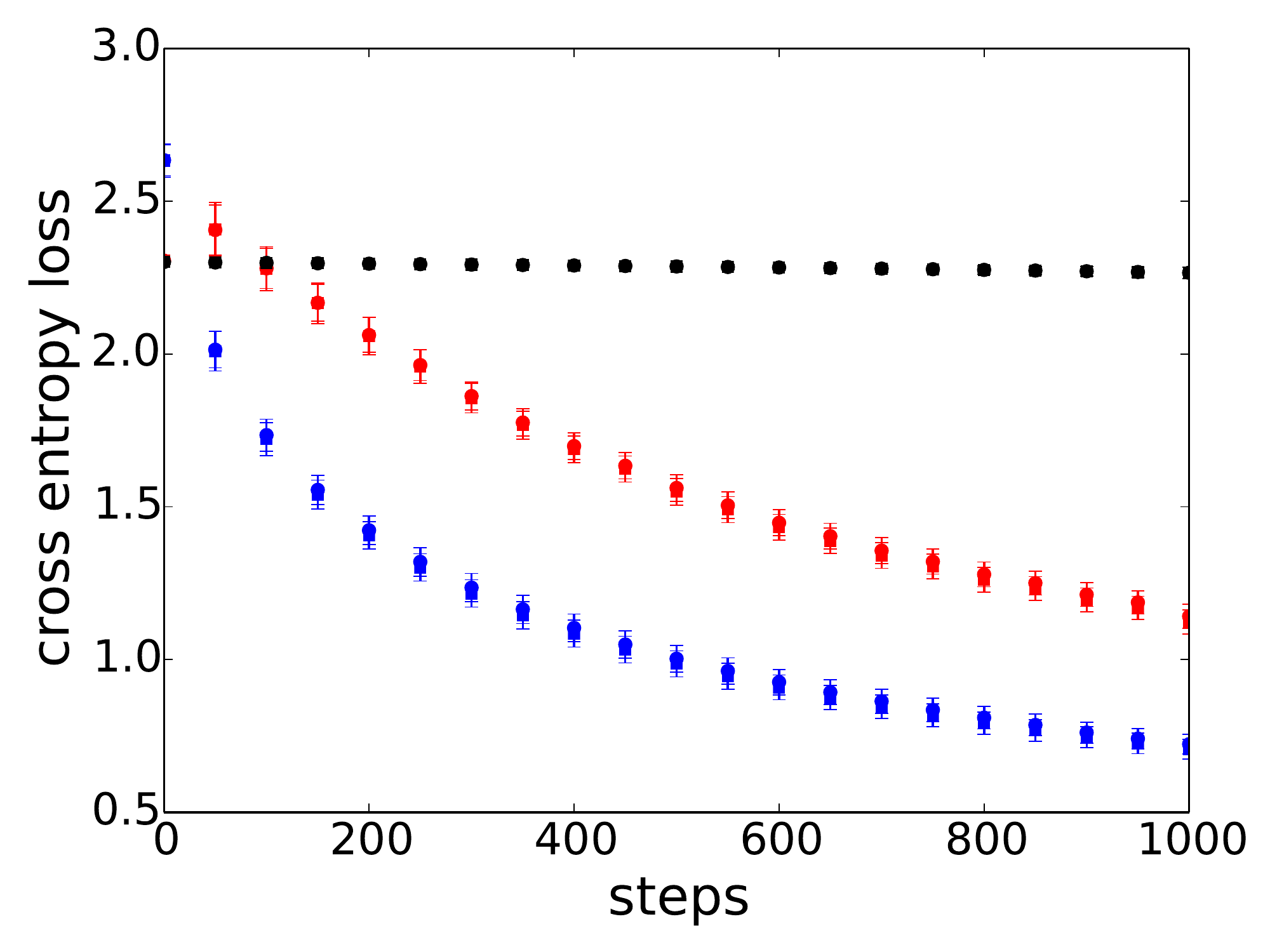} \\
\includegraphics[width=0.75\linewidth]{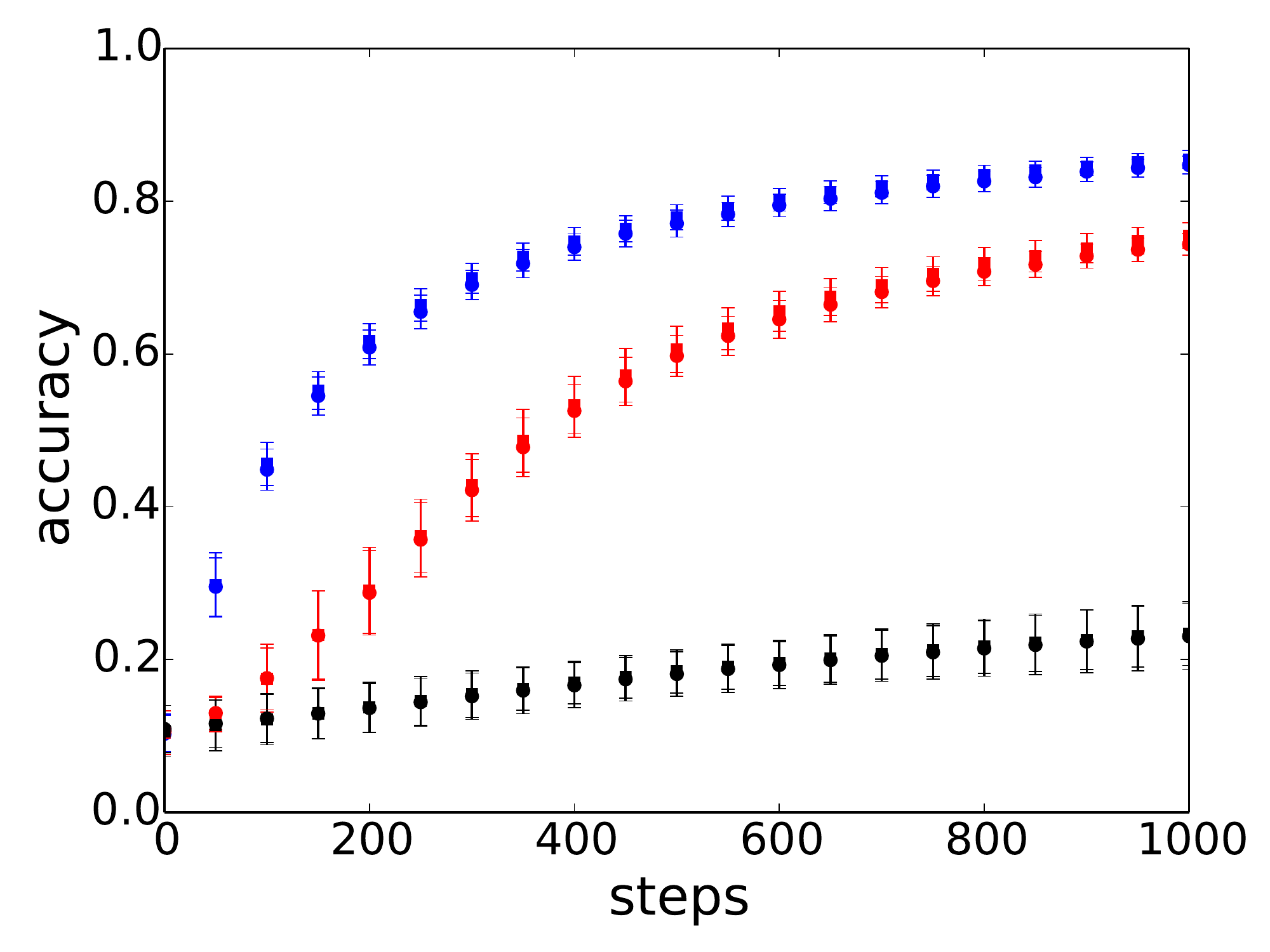}
\end{center}
\caption{\label{fig:mnist_cnn} Standardization loss accelerates convergence for 3-layer CNN trained on MNIST. (top) Cross entropy loss and (bottom) accuracy across 1000 gradient steps at a low learning rate in a baseline model without normalization (black), standardization loss (red), and batch normalization network (blue) for training (circles) and validation (squares) data. Error bars represent standard deviation across 10 runs.}
\end{figure}

\begin{figure}[t]
\begin{center}
\includegraphics[width=0.75\linewidth]{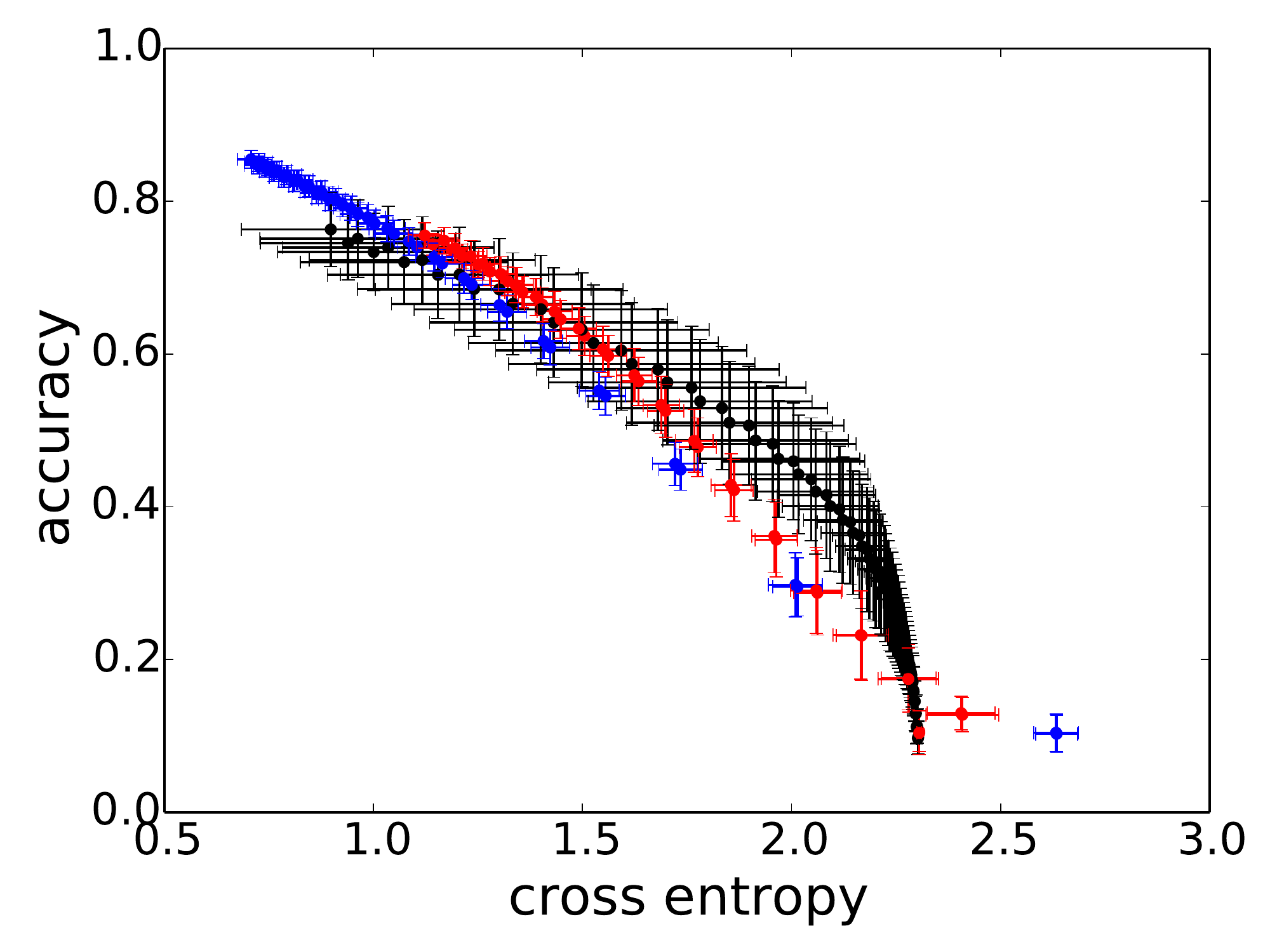} \\
\includegraphics[width=0.75\linewidth]{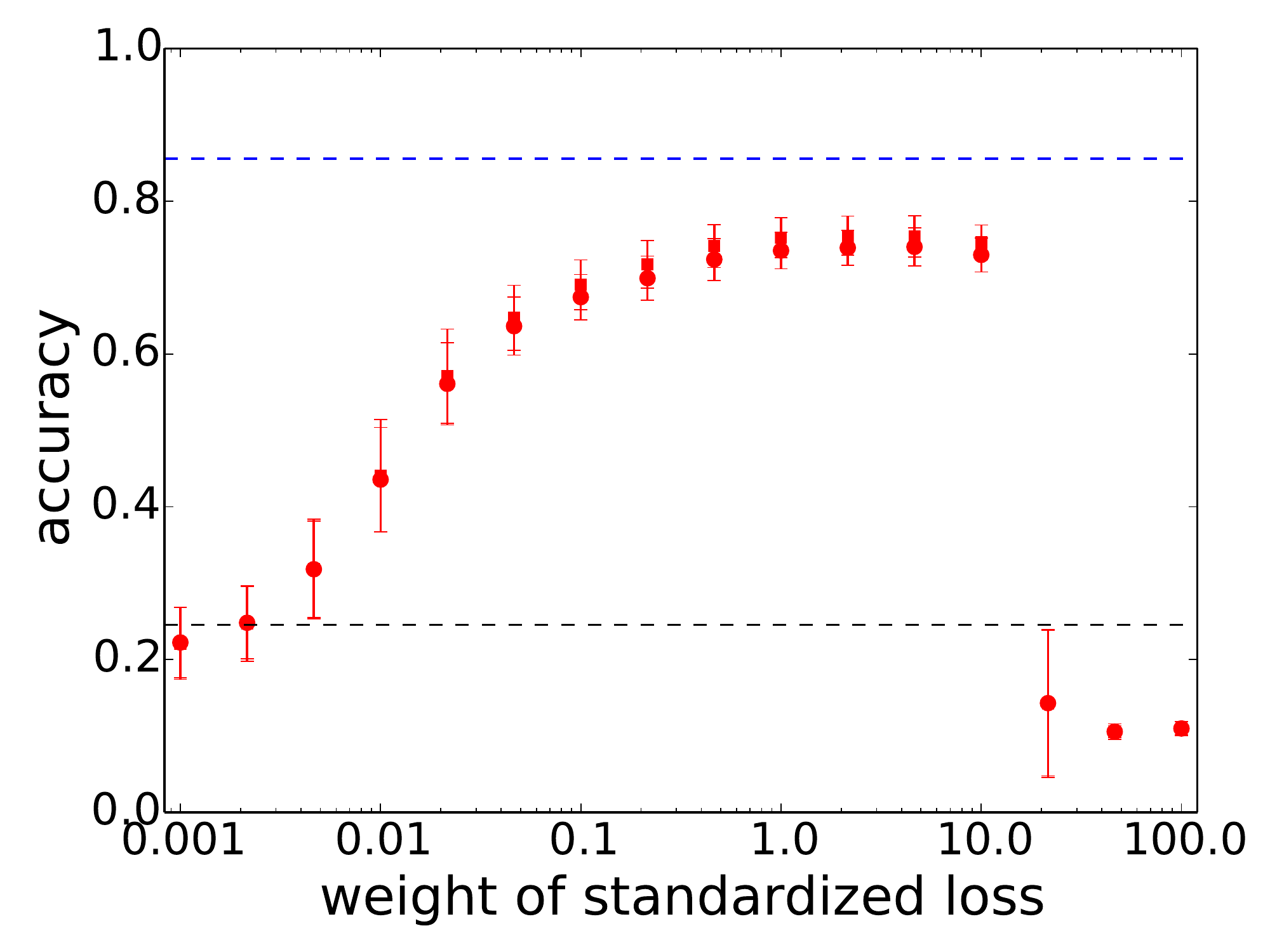}
\end{center}
\caption{\label{fig:mnist_robustness} Regularization and robustness of standardization loss in a 3-layer CNN.
(top) Cross entropy loss vs accuracy for MNIST. Note that the baseline network must be trained for 5000 steps in order to achieve comparable accuracies. (bottom) Accuracy after 1000 gradient steps across varying weight on standardization loss. Colors and marks follow Figure \ref{fig:mnist_cnn}.}
\end{figure}

We first examined the standardization loss in a small-scale experimental setting.
Specifically, we asked how the simple application of a standardization loss fares as a drop-in replacement for BN.
\footnote{In this experiment and in general we compare methods while keeping training hyperparameters identical. This is done to simplify experiments as well as demonstrate that our method can act as a good drop-in replacement for batch normalization.}
The goal of these experiments is to measure the predictive performance that different normalization configurations attain given a limited and constrained training budget (1000 SGD training steps at a low learning rate of 1e-3).

We constructed a small 3-layer CNN (see Appendix for details) and report the mean and standard deviation of the cross entropy loss and cross-validated accuracy across 10 training runs (Figure \ref{fig:mnist_cnn}). 
With BN, the CNN architecture achieves a reasonable accuracy (85.6\% $\pm$ $1.1\%$) with a small budget.
As expected, removing BN from this network drastically reduced predictive performance (24.6\% $\pm$ 6.5\%) given an identical, limited training regiment.
However, applying an additional standardization loss (weight $= 1.0$) to the activations resulted in a performance of 75.2\% $\pm$ 2.6\%, restoring most of the performance drop due to removing BN.

Across the 1000 training steps, we observed that the application of a standardization loss reliably accelerated training on the primary classification objective for both training and evaluation data, compared to the baseline model without normalization. (Figure \ref{fig:mnist_cnn}, top, red vs black). In addition, the application of the standardization loss accelerated arriving at higher predictive accuracies (bottom, red vs black). Note that in all cases the standardization loss did not accelerate training as fast as BN (blue).

We next asked whether the gain in performance from adding the standardization loss was due to a regularization effect. For example, weight decay is known to improve cross-validated accuracy but does not necessarily accelerate the convergence \citep{goodfellow2016deep, murphy2013ml}. Figure \ref{fig:mnist_robustness} (top) measures the cross entropy loss vs the accuracy for the baseline model (black), BN (blue) and standardization loss (red). If the standardization loss acted as a regularizer that improves the cross-validated accuracy, then the red points would be shifted upward indicating better predictive performance given the same cross entropy value. We do observe a slight upward shift with respect to BN but not the baseline model. We take these results to indicate that most of the gains of the standardization loss over the baseline is instead due to accelerated training.

When using the standardization loss, the total loss becomes a combination of the standardization loss and the primary objective. To explore how sensitive the model is to the weight attached to the standardization loss, we systematically varied the strength of the loss and measured the performance of the model after 1000 training steps with the same, limited training budget (Figure \ref{fig:mnist_robustness}, bottom). We found that increasing the standardization loss over four orders of magnitude [0.001, 10.0] increased the cross-validated performance -- although beyond a particular weighting the predictive performance collapsed.

\begin{figure}
\begin{center}
\includegraphics[width=0.75\linewidth]{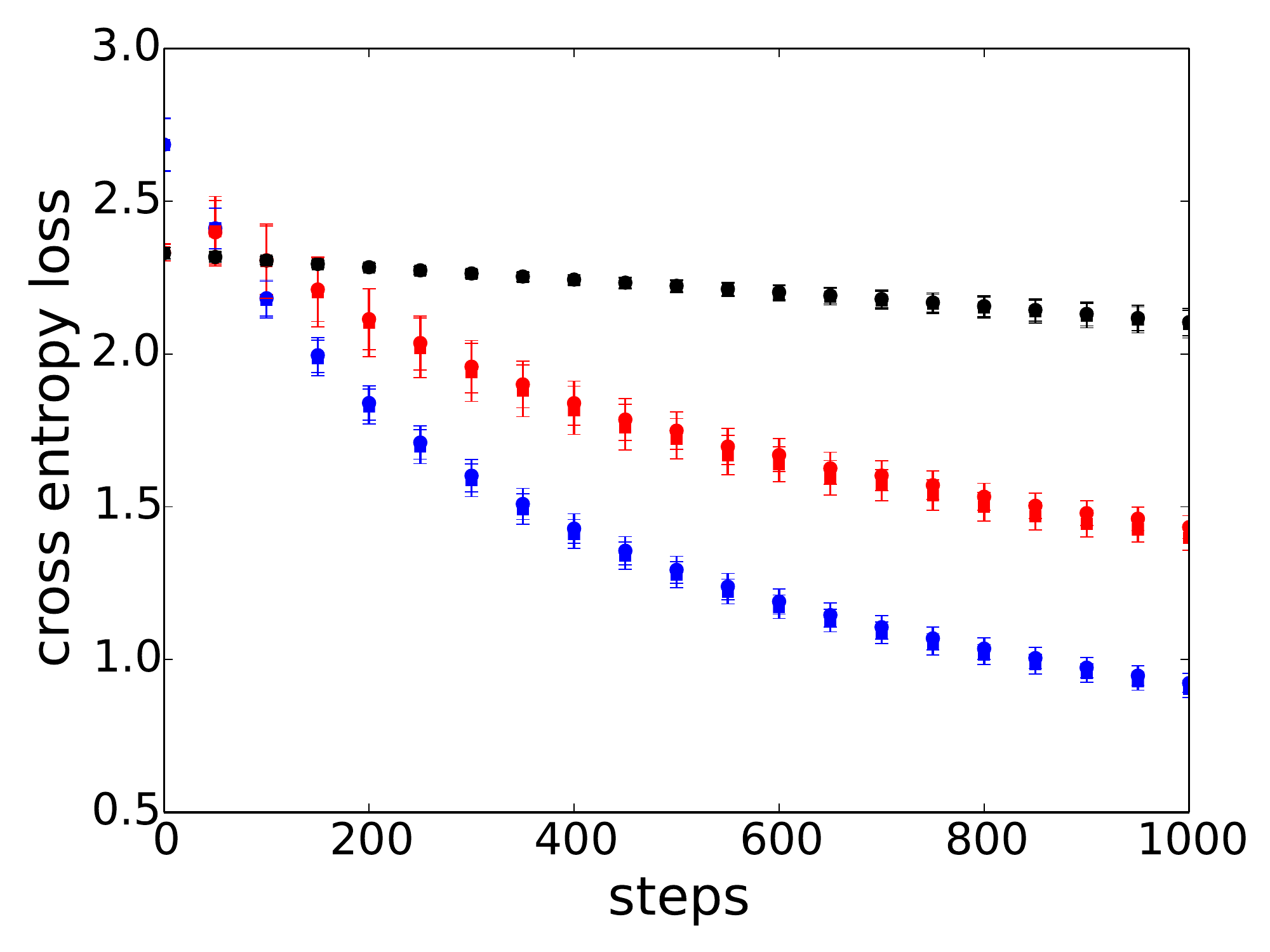} \\
\includegraphics[width=0.75\linewidth]{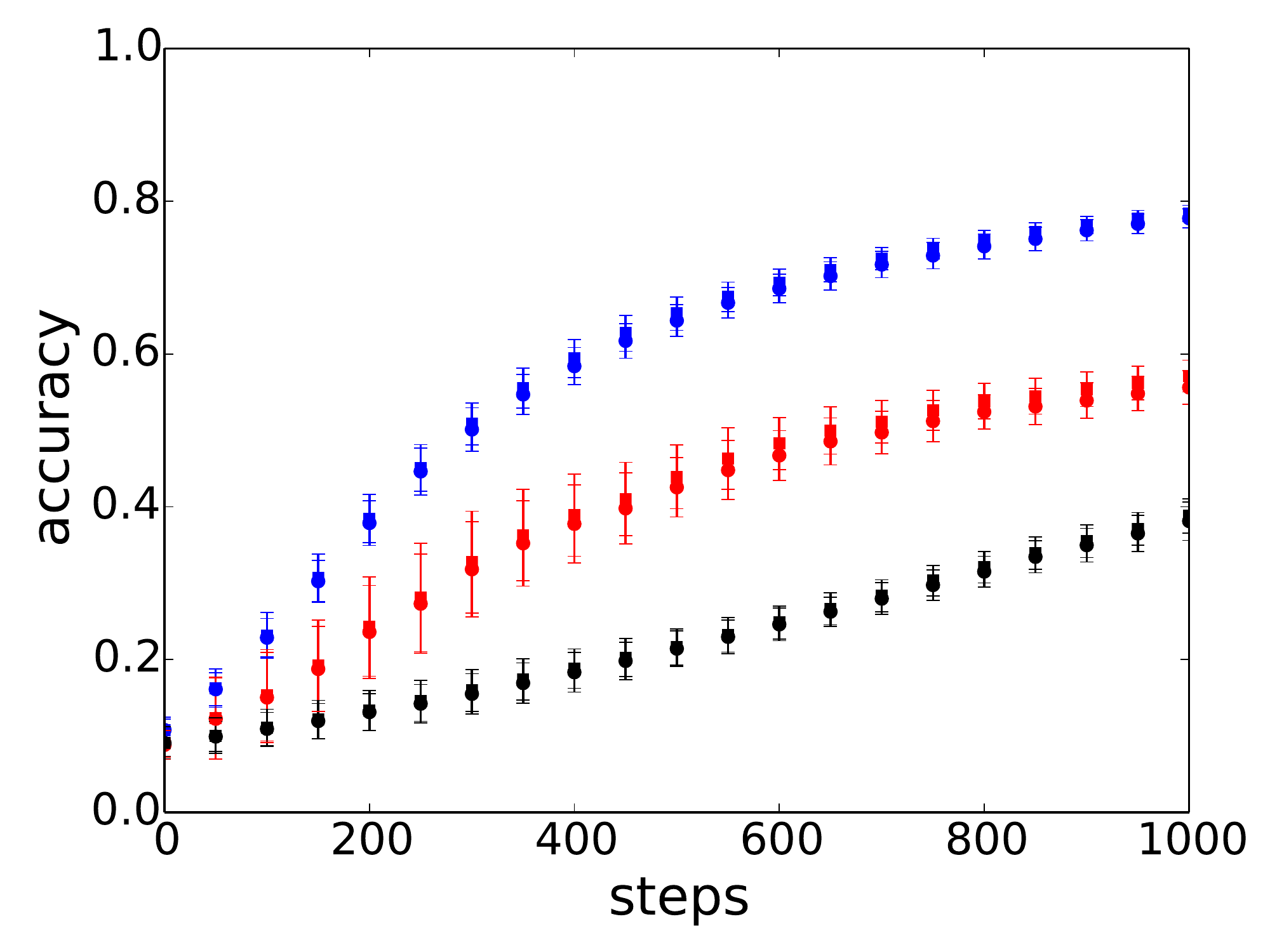}
\end{center}
\caption{\label{fig:mnist_mlp} Standardization loss accelerates convergence for 3-layer MLP trained on MNIST. Panels and colors follow Figure \ref{fig:mnist_cnn}.}
\end{figure}

To explore the generality of our approach, we also examined how well a standardization loss could improve the training performance in a small-scale MLP.
We constructed a 3-layer MLP and trained the model on MNIST using the same limited training budget as in the CNN experiments.
Figure \ref{fig:mnist_mlp} demonstrates that the application of standardization loss accelerates training both in terms of achieving a lower primary objective (cross entropy) and an improved cross-validated accuracy (red vs black) even though the architectures are identical. Parallel to the CNN version, we do find that although a standardization loss accelerates training, it does not accelerate training as fast as BN (blue). Overall, we took these results on a toy example as a positive signal that the simple application of a secondary standardization loss may accelerate the training on the primary objective of a network \footnote{We did not pursue achieving state-of-the-art on MNIST because we view the performance as saturated given an arbitrary budget of training steps and hyperparameter tuning. For instance, all network configurations explored in this study could exceed 98\% cross-validated accuracy on MNIST given the freedom to increase the training budget and explore hyperparameter settings.}.

\begin{figure*}[t]
\begin{center}
\includegraphics[width=0.3\linewidth]{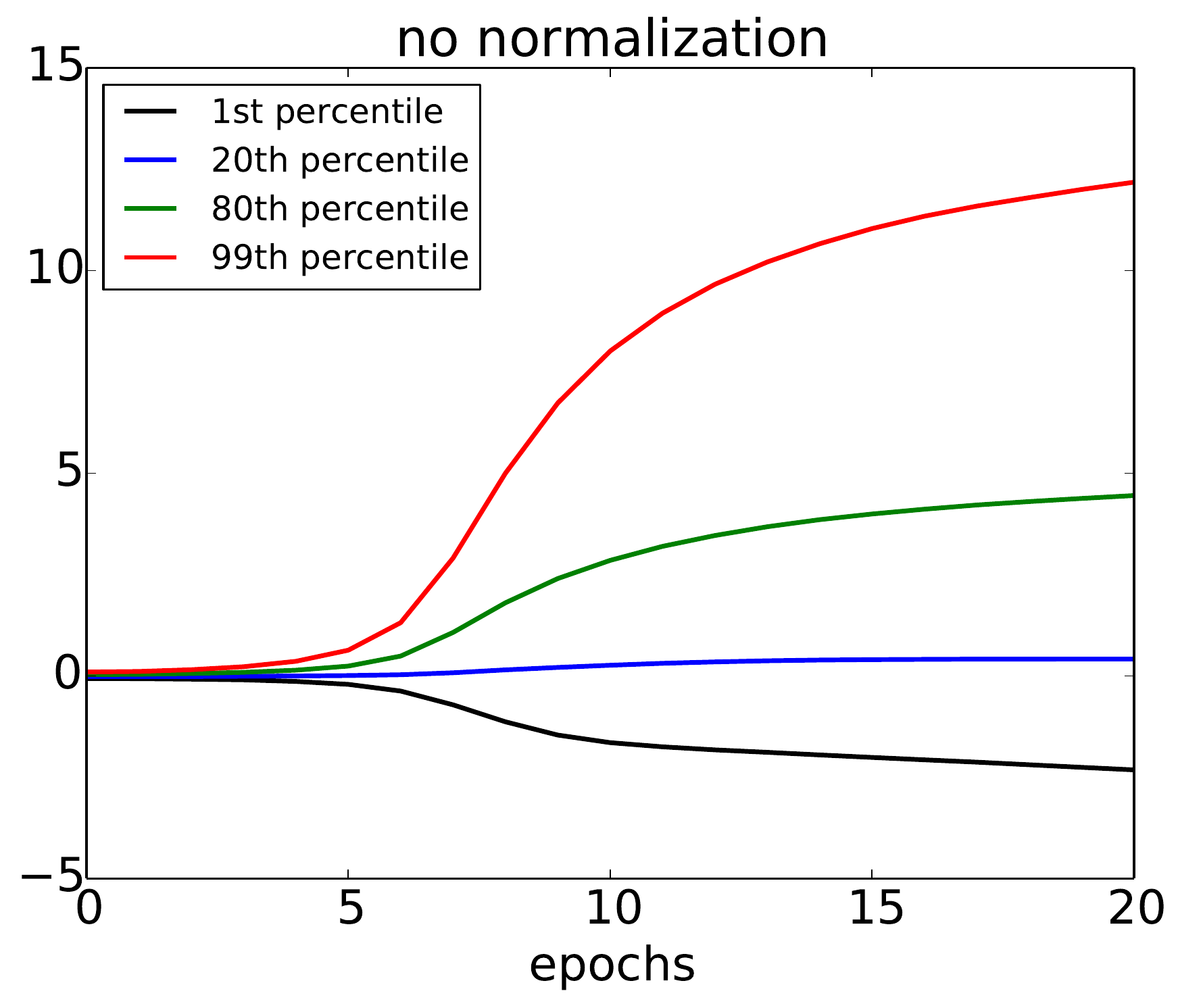}
\includegraphics[width=0.3\linewidth]{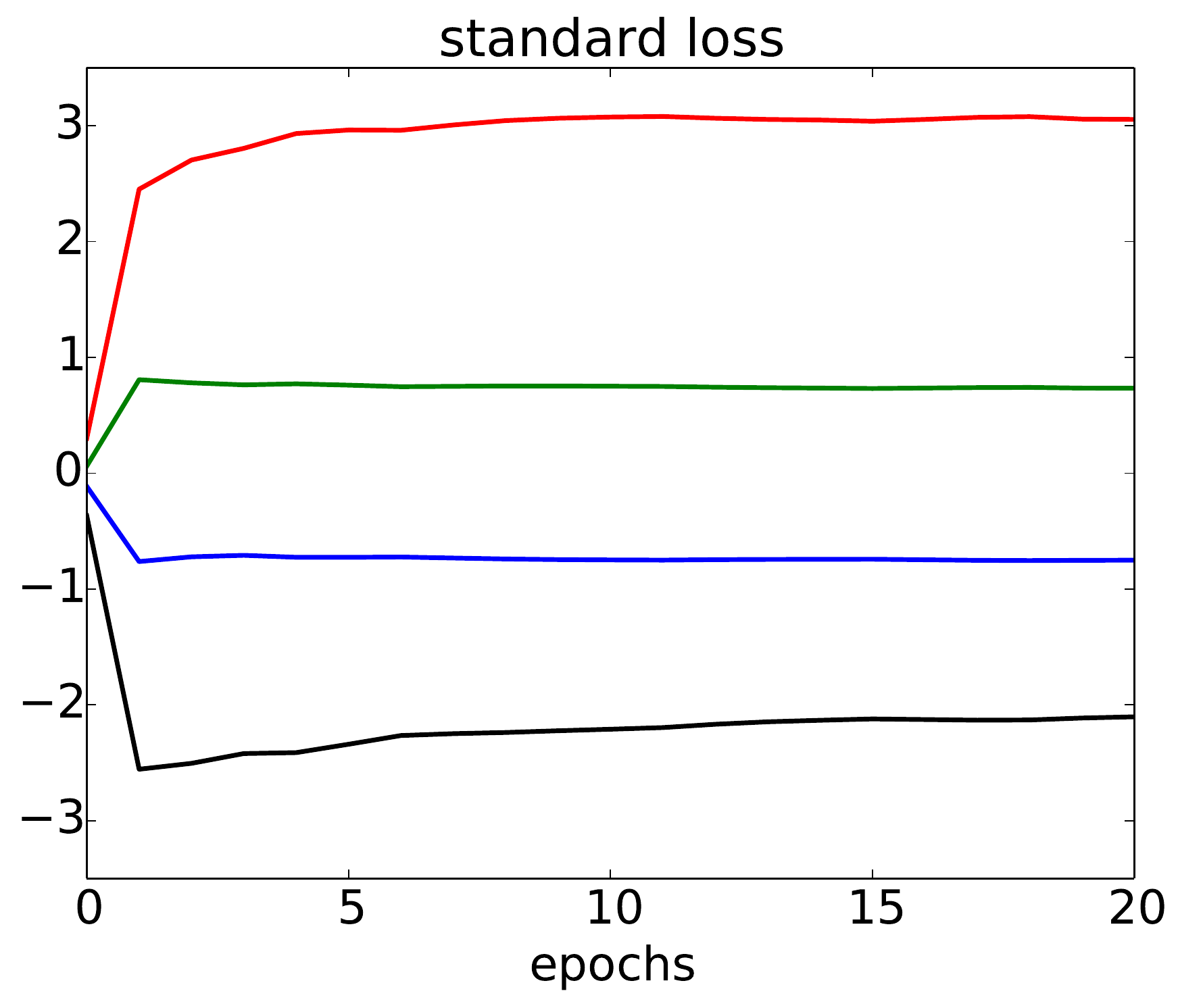}
\includegraphics[width=0.3\linewidth]{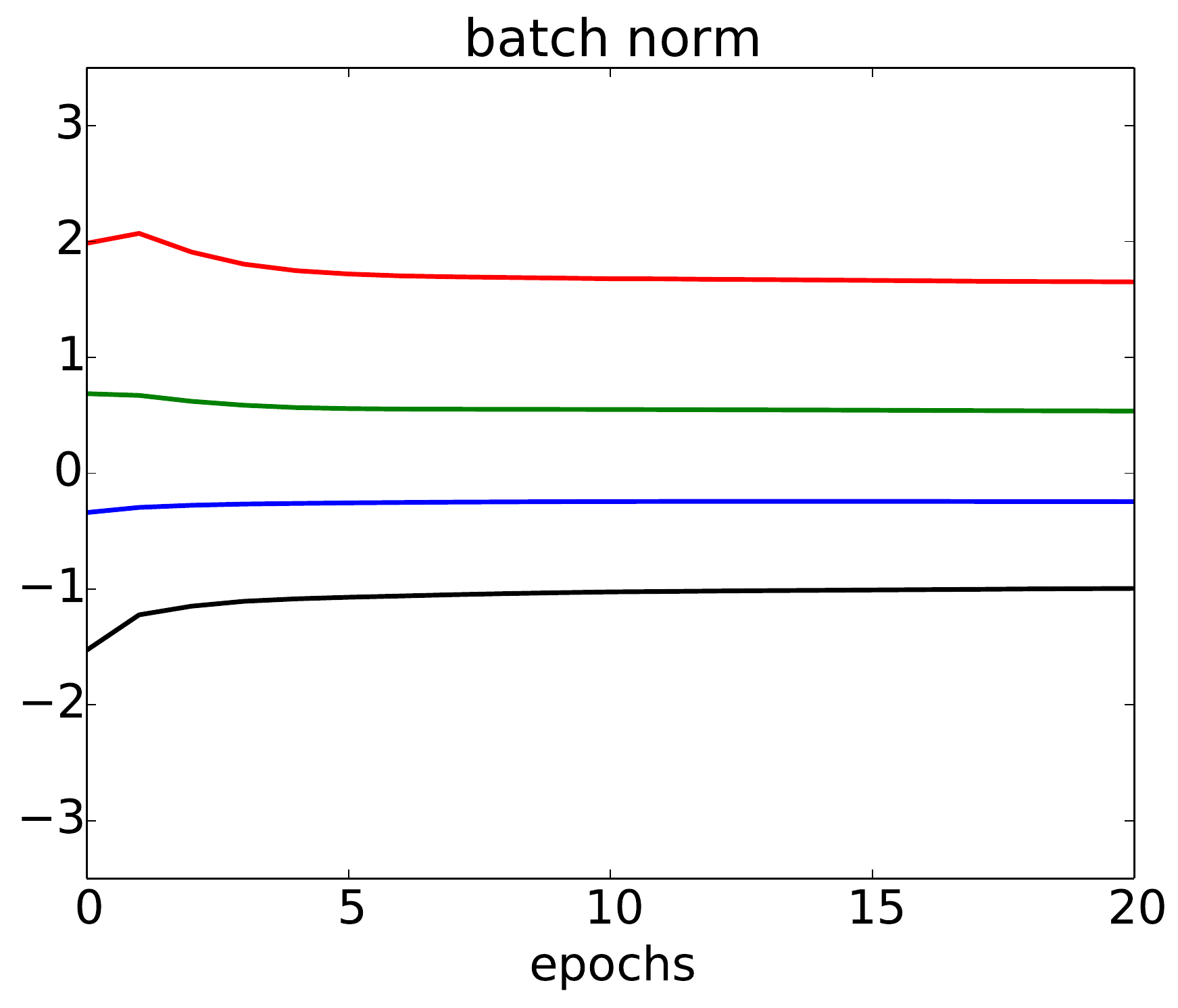}
\end{center}
\caption{\label{fig:dist}Evolution of CNN feature distributions for no normalization, standardization loss, and batch normalization trained on MNIST. Black, blue, green and red curves measure the \nth{1}, \nth{20}, \nth{80}, and \nth{99} percentile of the cumulative distributions for the final convolutional layer across training steps. Note the different scale of the y-axis for the no normalization plot.}
\end{figure*}

\subsection{Standardization loss results in a similar evolution of feature distributions}
Maintaining stability in the distribution of network features has been a motivating factor for introducing normalization into network architectures \citep{BatchNorm} (but see \cite{santurkar2018does,bjorck2018understanding,kohler2018towards}). We next examine how the feature distributions of a layer evolve across training steps using the 3-layer CNN from our previous experiments.
We train the network for 20 epochs (roughly 8.6K steps) and track the distribution of the final layer's activation across training steps (Figure \ref{fig:dist}).

Consistent with previous observations \citep{BatchNorm,wu2018group}, removing normalization from the network leads to activation distributions that change notably in shape and scale over the course of training (Figure \ref{fig:dist}, left). Applying normalization either through BN or a standardization loss restores control over the evolution of the feature distributions (Figure \ref{fig:dist}, middle and right). 
Importantly, although the network with the standardization loss is identical in architecture to the baseline network, we see that the simple application of a standardization loss results in a stable distribution that evolves in a way qualitatively similar to a technique which explicitly enforces normalization.

\subsection{Standardization loss accelerates training of large-scale MLPs}

\begin{figure}[t]
\begin{center}
\includegraphics[width=0.8\linewidth]{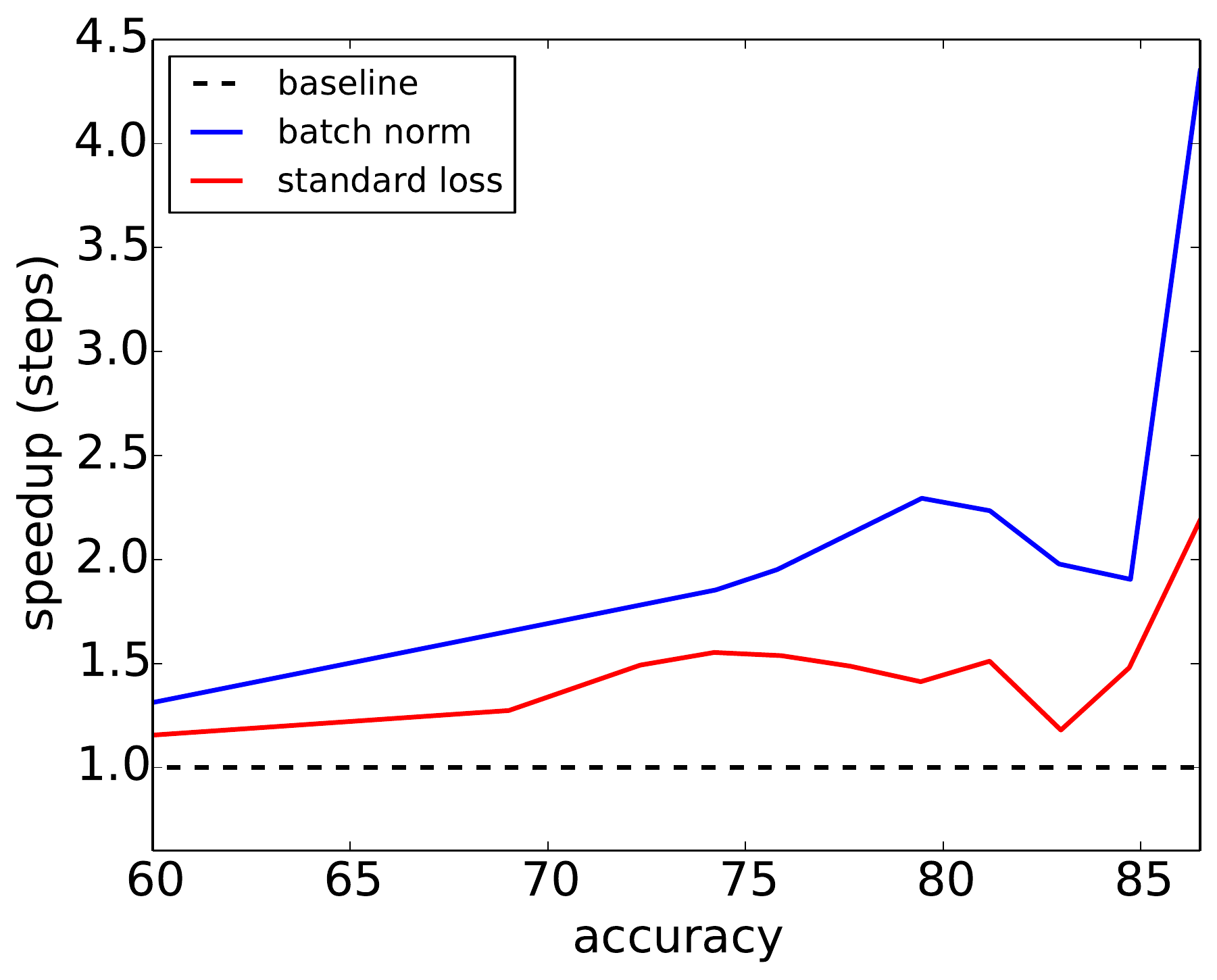} \\
\vspace{0.25cm}
\begin{tabular}{l|c|c}
& \# steps & speed-up \\
\hline
 baseline & 26.6K & -  \\ 
 batch norm & 6.1K & 4.4\,x \\  
 standard loss & 12.1K & 2.2\,x \\ 
\end{tabular}
\end{center}
\caption{Standardization loss accelerates large-batch training in an MLP on ModelNet40. 
Speed-ups in table reflect number of steps to reach the maximum accuracy (86.5\%) achieved by the baseline model without normalization. See text for details.\label{fig:mlp-pointnet}}
\end{figure}

After observing positive results for MLPs trained with standardization loss on MNIST, we next moved on to a large-scale task for MLP networks. Classification of point cloud data remains a challenging learning task with applications to robotics and self-driving cars. One common benchmark for assessing performance on point cloud data is ModelNet-40 \citep{wu20153d} -- a classification dataset consisting of 12,311 CAD models
across 40 man-made categories. Most state-of-the-art methods on point cloud data employ an MLP with BN as a central component of the architecture \citep{qi2017pointnet,qi2017pointnetplusplus}.

We focus on PointNet \citep{qi2017pointnet} as a large-scale test of a standardization loss. Our implementation of PointNet with BN achieves 89.3\% accuracy, compared to a published value of 89.2\%. We also remove BN from PointNet to measure the performance of a no normalization baseline, and find the resulting model achieves an accuracy of 86.5\%.

We measured the acceleration of the normalization models over the baseline by calculating the ratio of the number of training steps required to achieve a given accuracy, for the baseline model compared to the normalization model. For instance, the baseline model requires 26.6K training steps to achieve its maximum accuracy (86.5\%). The equivalent model with BN only requires 6.1K steps to achieve this same accuracy. Hence, we measure a speed-up due to BN of 4.4\,x ($=26.6 / 6.1$). Figure \ref{fig:mlp-pointnet} (blue) plots this speed-up across the entire course of training for BN indicating that BN is consistently faster than no normalization.

We next add a standardization loss to the baseline model, using an identical training scheme (Figure \ref{fig:mlp-pointnet}, red). The resulting model achieves a 2.2\,x speed-up and likewise demonstrates accelerated accuracies over the baseline model across the entire training session. 
Again we see that a simple secondary objective can achieve some of the gains of BN on a real world task without any of the complications.

\subsection{Standardization loss accelerates training of large-scale CNNs}

\begin{figure}[t]
\begin{center}
\includegraphics[width=0.8\linewidth]{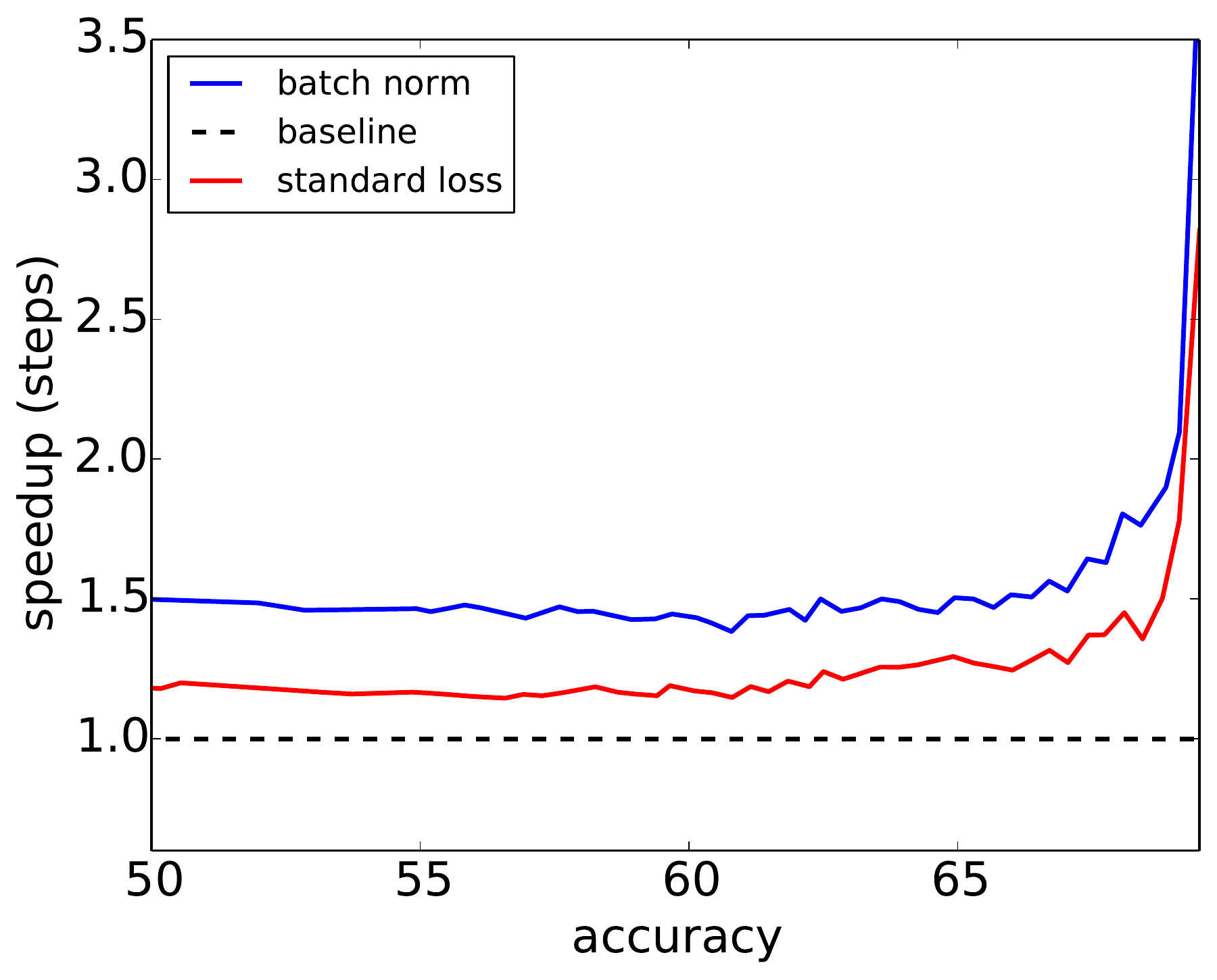}
\end{center}
\caption{\label{fig:imagenet-inception-v2}Acceleration of ImageNet top-1 validation accuracy on Inception-v2 over a baseline network by the application of a standardization loss (red) or batch normalization (blue).
See Table \ref{table:imagenet-acceleration} for summary. 
}
\end{figure}

\begin{table}
\begin{center}
\begin{tabular}{l|c|c}
Inception-v2 & \# steps & speed-up \\
\hline
 baseline & 1972K & -  \\ 
 batch norm & 534K & 3.7\,x \\  
 standard loss & 700K & 2.8\,x \\ 
\end{tabular} \\
\vspace{0.4cm}
\begin{tabular}{l|c|c}
ResNet-50 & \# steps & speed-up \\
\hline
 baseline & 89.4K & -  \\ 
 batch norm & 39.5K & 2.3\,x \\  
 standard loss & 40.0K & 2.2\,x \\ 
\end{tabular}
\end{center}
\caption{\label{table:imagenet-acceleration}Acceleration of ImageNet training on Inception-v2 and ResNet-50. Comparison of baseline model with batch normalization and standardization loss. All speed-ups measured in terms of the number of steps to reach the maximum accuracy achieved by the baseline model (69.5\% and 70\% accuracy, respectively). See text for details.}
\end{table}

Given consistent results for both small- and large-scale MLPs, we next asked if there was an analogous result for CNNs.
We explored this question in the context of  
the ImageNet dataset \citep{deng2009imagenet},
across three large-scale CNN architectures:  Inception-v2 \citep{szegedy2016rethinking}, ResNet-50 \citep{he2015deep}, and MobileNet-v1 \citep{howard2017mobilenets}.

We examined the ResNet-50 and Inception-v2 architectures in order to measure the speed-ups due to normalization. As with PointNet, both CNNs employ BN as a default and thus their hyperparameters are tuned to the use of BN. We found that in order to stabilize Inception-v2 for the baseline setting, it was necessary to lower the learning rate; thus, all comparisons for Inception-v2 are made at this low learning rate (see Appendix for details). Figure \ref{fig:imagenet-inception-v2} measures the speed-up gains over a baseline network by applying the standardization loss (red) and using BN (blue) over the course of training. 
The standardization loss achieves speed-ups comparable to BN of 2.8\,x vs 3.7\,x (Table \ref{table:imagenet-acceleration}, top).

To further explore the generality of the results, we also measure the acceleration of ResNet-50 trained on ImageNet. The baseline model proved more robust to a lack of normalization and did not require adjusting the default learning rate. In line with Inception-v2, we find that the application of a standardization loss accelerates training comparable to BN, with speed-ups of 2.2\,x vs 2.3\,x (Table \ref{table:imagenet-acceleration}, bottom).

\subsection{Models trained with a standardization loss are robust across batch sizes}

BN is known to suffer in performance when trained with very small batch sizes. This issue is known to arise due to the rigid requirement that every batch must be normalized -- even when the moments of a distribution are poorly estimated as is the case in a small batch size. In contrast, a standardization loss merely promotes 
normalization and thus may be more resilient to poorly estimated moments in individual batches. This intuition motivates us to test if indeed a standardization loss may be more robust to choice of batch size and alleviate this major drawback of BN.

We consider the 
MobileNet-v1 architecture as
this architecture is often employed in domains
where batch size is restrictive. Similar to the previous networks we explored, MobileNet-v1 has BN built-in to its architecture. Interestingly, we found that removing BN from MobileNet-v1 destroyed the ability of the network to train above chance rate, even after tuning across a broad range of reasonable hyperparameters.
Conversely, by the simple application of the standardization loss (with identical training hyperparameters), MobileNet-v1's predictive accuracy was largely restored (70.6\% vs 66.6\% top-1 accuracy at batch size 32).

\begin{figure}[t]
\begin{center}
\includegraphics[width=0.8\linewidth]{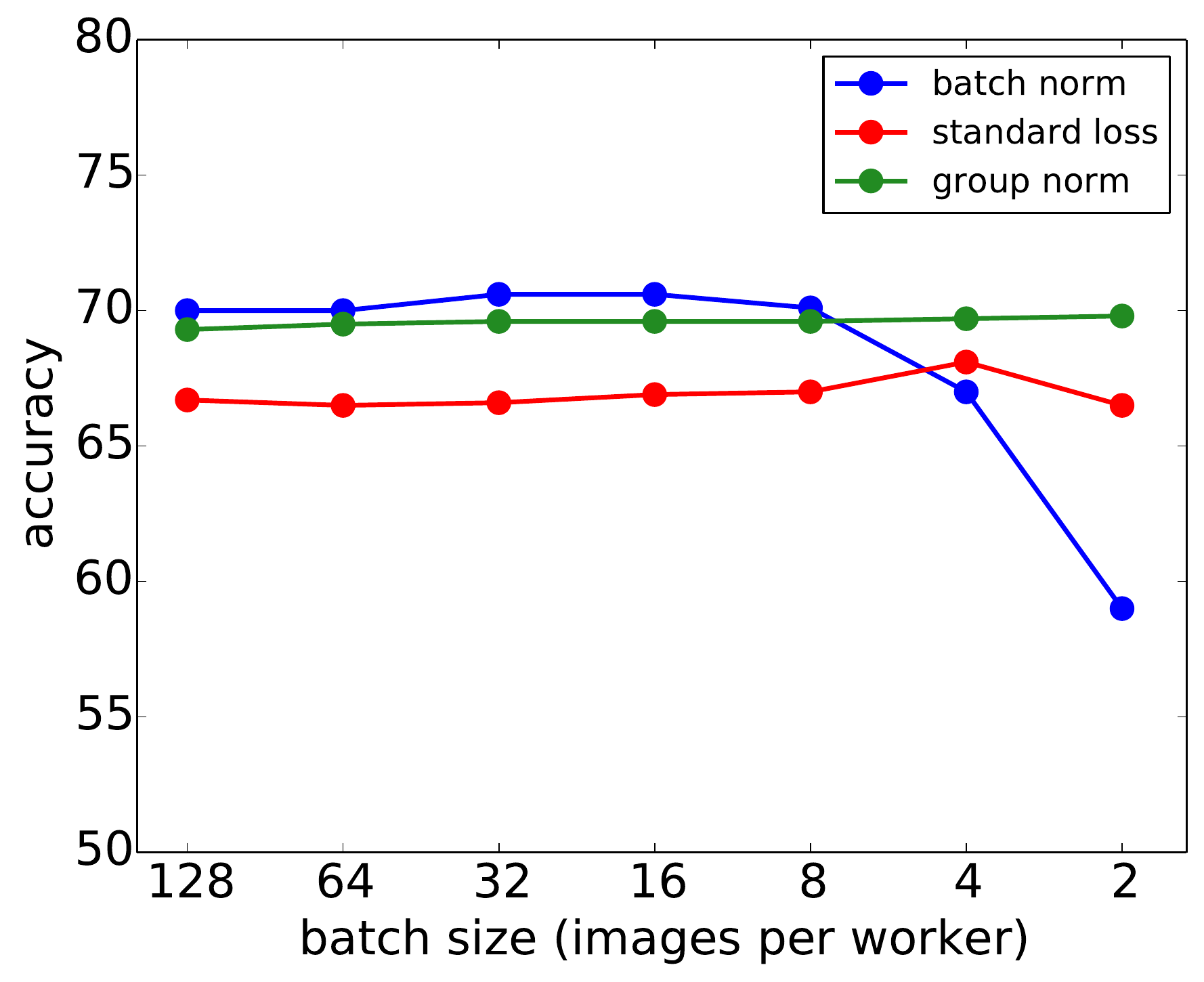}
\end{center}
\caption{\label{fig:mobilenet-smallbatch}Top-1 validation accuracy of batch normalization, group normalization, and standardization loss networks trained with different batch sizes.}
\end{figure}

To investigate the effect of small batch size, we trained MobileNet-v1 ranging in batch sizes from 2 to 128, adopting a linear scaling rule for adjusting the learning rate \citep{goyal2018accurate}. We compared the performance of BN to the standardization loss across batch size (Figure \ref{fig:mobilenet-smallbatch}, blue vs red). As previously reported, the performance of BN varies notably across different batch sizes -- with small batch sizes typically leading to worse performance. In contrast, a network trained with a standardization loss achieves roughly constant performance (66.5\% - 66.9\%) across the wide range of batch sizes and demonstrates no precipitous drop in performance at small batch sizes as observed with BN. For batch sizes 2 and 4, the standardization loss outperforms BN.

As previously mentioned, other techniques have recently been introduced to specifically address the issue of the poor small batch performance of BN. Batch renormalization \citep{ioffe2017batch} mitigates this variability to a degree, but still exhibits some batch dependencies. Group normalization (GN) \citep{wu2018group} offers a method for calculating normalization statistics that are not dependent on the batch dimension, and achieves state-of-the-art performance for CNNs trained with small batch sizes. As GN represents the state-of-the-art in small batch training of image models, we additionally chose to compare the performance of the standardization loss against GN in MobileNet-v1 training across a large range of batch sizes
\footnote{Although the primary goal of GN was asymptotic accuracy rather than training acceleration, 
we found GN to likewise accelerate Inception-v2 and ResNet-50 training at rates comparable to a standardization loss.}.
Figure \ref{fig:mobilenet-smallbatch} (green) highlights that GN indeed exhibits robustness to batch size and outperforms the standardization loss.
Note that GN employs a distinct and complimentary form of normalization that is specific to CNNs and there are trade-offs in using these techniques that are addressed in the Discussion. Nonetheless, the application of a standardization loss does provide a simple method for accelerating training that is largely invariant to batch size.

\section{Discussion}

In this work we have introduced a simple auxiliary loss that encourages the distribution of activations in a deep network towards a standardized Gaussian. We find that adding this secondary objective accelerates the training by almost 2-fold on the primary classification objective for a range of network architectures, and the resulting network can be trained across a wide range of batch sizes. 

The gain of our method over other normalization techniques is simplicity and wide applicability to several architectures and problems. Existing normalization techniques are highly specialized and only apply to certain neural network architectures and training setups. BN performs poorly when the activation moments are poorly estimated (e.g. small batch size, non-{\it i.i.d.} batches). Additionally, BN requires the user to maintain moving averages at test-time which makes it difficult to apply to certain architectures (e.g. recurrent neural networks). GN addresses the small batch size problem of BN, but it is only applicable to convolutional models, and therefore another technique must be used for training MLPs and recurrent networks with small or non-{\it i.i.d.}. batches (e.g. layer normalization, LN). LN empirically works well for MLPs but often does not perform as well as BN for CNNs \citep{ba2016layer}. Further, techniques like GN and LN require continued normalization at test-time, demanding an extra pass over the activations at each layer of the network for each prediction step. This may be prohibitive for memory-constrained devices (e.g. mobile and edge devices) and may slow down inference notably. Weight normalization is a simple and architecture-invariant approach that is very successful for small-scale problems, but does not seem to scale well to large-scale problems \cite{gitman2017comparison}. In the experimental settings we considered, our approach to normalization is (1) robust to batch size, (2) applies to both MLPs and CNNs, (3) does not require explicit inference-time normalization, and (4) works in both small- and large-scale settings. 

While we find that our method is successful in accelerating training and asymptotic accuracy over networks without normalization, we see lower asymptotic accuracies compared to BN. We note that we did not re-tune hyperparameters for models built with BN as a default, and it is likely that re-tuning could give an accuracy boost. Another possible explanation for the performance difference is that explicit normalization techniques ensure standardized activations from the very first training step, whereas training with a standardization loss requires multiple steps of SGD before the loss decreases and the activations are sufficiently standardized. 
Better understanding of the differences between a loss-based approach to normalization and existing explicit normalization techniques may allow for networks to benefit even more from a standardization loss.

This work opens up several interesting avenues for further exploration of the application of a standardization loss. 
Recurrent networks present a challenging domain in which standard normalization techniques do not work well, and specialized variants are required \citep{ba2016layer, laurent2016batch, cooijmans2016recurrent}. Because a standardization loss provides a less stringent form of normalization, it would be interesting to explore the degree to which this method generalizes to these forms of architectures.
Likewise, the application of a standardization loss may be particularly advantageous in a stochastic training setting in which the data is not {\it i.i.d.} (e.g. metric learning, continual learning) and enforcing that a given batch should be strictly normalized is overly restrictive. Commonly used normalization methods are impeded because the batch statistics provide a skewed perspective of the dataset statistics, which in turn limits training performance (but see \cite{ioffe2017batch}).

This work focused on encouraging the activations of a given layer towards a standardized Gaussian distribution. Another interesting direction is to apply the same standardization loss but encourage other types of distributional forms. For instance, a simple extension is to require that the activations at a given layer are not just standardized but decorrelated as well. Although more computationally demanding, such a requirement may lead to accuracy improvements  \citep{huang2018decorrelated}. Another direction to consider is sparsity. Sparsity is well known to improve generalization, and replacing the target distribution with a Laplace or Bernoulli distribution may yield benefits in representational ability, particularly in the context of unsupervised learning \citep{lee2008sparse,nair20093d,gao2016group,ngiam2011sparse}.
\newpage

\newpage
\bibliography{paper}
\bibliographystyle{icml2019}

\clearpage
\appendix

\section{Details of model training}

\subsection{CNN MNIST}
We trained a small 3-layer CNN with 32, 16 and 8 filters ($5\times5$, stride 2) in each layer and ReLU nonlinearities. The network was trained with SGD at a learning rate of 1e-3 and a batch size of 128. We trained the model for 1000  gradient  updates, which corresponds to training with 2.13 epochs of the training set. All models used a weight decay of 4e-5 and the network trained with standardization loss used a weight of 1.0. The BN network used an additive shift parameter after the normalization at each layer (no scale parameter was necessary due to the use of the ReLU nonlinearity). 

\subsection{MLP MNIST}
The MLP we trained had 3-layers with 100 units per layer and used ReLU nonlinearities. We used an identical training scheme to the CNN training. That is, we trained for 1000 SGD steps with a batch size of 128, learning rate of 1e-3 and weight decay of 4e-5. The standardization loss was weighted by 1.0, and the BN network had shift parameters at each layer.

\subsection{PointNet \citep{qi2017pointnet}}
We trained a PointNet with no normalization, standardization loss, and BN each on a single GPU. For all experiments PointNet was optimized using the ADAM optimizer with momentum of 0.9, and a starting learning rate of 1e-3 that was decayed by half every 20 epochs. We used a batch size of 32 and ReLU nonlinearities. The BN network applied a scale and shift at each layer. We also found that the standardization loss network benefited from such scale and shift parameters. The standardization loss is weighted 1e-3.

\subsection{Inception-v2  \citep{szegedy2016rethinking}}
We trained Inception-v2 on 8 Tensor Processing Unit (TPU) cores at a batch size of 128 images per core (1024 total). We used the RMSProp optimizer with a base learning rate of 0.006 which decayed by 0.98 every 12 epochs. The model uses the ReLU nonlinearity. The BN architecture adds a shift parameter after every normalization operation. We used a standardization loss weight of 1e-5.

\subsection{ResNet-50 \citep{he2015deep}}
ResNet-50 was trained on 8 TPU cores at a batch size of 128 examples per core, using Nesterov momentum with a momentum coefficient of 0.9. The learning rate schedule had a gradual warmup phase, starting at 0 and increasing linearly per step for the first 5 epochs to a maximum learning rate of 0.1. After this, it was decayed by 0.9 after 30, 60, and 80 training epochs. BN added a shift parameter after each explicit normalization. The standardization loss is weighted by 1e-6.

\subsection{MobileNet-v1 \citep{howard2017mobilenets}}
We trained a MobileNet on 8 TPU cores, for each batch size (per core) in \{2, 4, 8, 16, 32, 64, 128\}. Total batch size is given by batch size per core $\times$ 8, but statistical moments are calculated independently for each core and are not shared across cores. We adopt the linear scaling rule from \citet{goyal2018accurate} for adjusting the learning rate to batch size, starting with a maximum learning rate of 0.165 at 128 images per core. The learning rate is decreased by 0.94 every 3 epochs and the RMSProp optimizer is used. The architecture uses ReLU6 and BN employs both scale and shift parameters after normalization at every layer. When using GN, we use a group size of 32 for all experiments. The standardization loss is weighted by 1e-4 for all experiments.

\end{document}